\documentclass[letterpaper, 10 pt, conference]{ieeeconf}  % Comment this line out if you need a4paper

\IEEEoverridecommandlockouts                              % This command is only needed if 
\usepackage{graphics} % for pdf, bitmapped graphics files
\usepackage{epsfig} % for postscript graphics files
\usepackage{amsmath} % assumes amsmath package installed
\usepackage{amssymb}  % assumes amsmath package installed
\usepackage{cite}
\makeatletter
\let\NAT@parse\undefined
\makeatother
\usepackage[backref, colorlinks, citecolor=blue]{hyperref}
\usepackage{soul}
\usepackage{booktabs}
\usepackage{multirow}

\title{\LARGE \bf
Rich State Observations Empower Reinforcement Learning to Surpass PID: A Drone Ball Balancing Study
}

\author{Mingjiang Liu and Hailong Huang% <-this % stops a space
\thanks{The authors are with the Autonomous and Interactive Mobile Systems Group (AIMS), Department of Aeronautical and Aviation Engineering, The Hong Kong Polytechnic University, Hong Kong, China. Email:
        {\tt\small mingjiangaae.liu@connect.polyu.hk}}%
\thanks{This paper has been accepted for presentation at the Advancements in Aerial Physical Interaction Workshop of the IEEE/RSJ International Conference on Intelligent Robots and Systems (IROS), 2025. Comments and feedback are welcome.}
}

\begin{document}

\maketitle
\thispagestyle{empty}
\pagestyle{empty}

%%%%%%%%%%%%%%%%%%%%%%%%%%%%%%%%%%%%%%%%%%%%%%%%%%%%%%%%%%%%%%%%%%%%%%%%%%%%%%%%
\begin{abstract}
This paper addresses a drone ball-balancing task, in which a drone stabilizes a ball atop a movable beam through cable-based interaction. We propose a hierarchical control framework that decouples high-level balancing policy from low-level drone control, and train a reinforcement learning (RL) policy to handle the high-level decision-making. Simulation results show that the RL policy achieves superior performance compared to carefully tuned PID controllers within the same hierarchical structure. Through systematic comparative analysis, we demonstrate that RL’s advantage stems not from improved parameter tuning or inherent nonlinear mapping capabilities, but from its ability to effectively utilize richer state observations. These findings underscore the critical role of comprehensive state representation in learning-based systems and suggest that enhanced sensing could be instrumental in improving controller performance.

\end{abstract}

% \begin{keywords}
%  Human-robot collaboration, reinforcement learning, and proportional-integral-derivative control.
% \end{keywords}

%%%%%%%%%%%%%%%%%%%%%%%%%%%%%%%%%%%%%%%%%%%%%%%%%%%%%%%%%%%%%%%%%%%%%%%%%%%%%%%%
\section{Introduction}

Aerial manipulation, which aims to endow aerial robots with the capability to physically interact with their surroundings \cite{2021_tro_aerial_manipulator}, has attracted significant attention in recent years. The most common aerial manipulation systems employ unmanned aerial vehicles (UAVs) as floating bases and equip them with robotic arms or custom grippers. These configurations have been successfully applied to tasks such as millimeter-level peg-in-hole insertion \cite{2023_TRO_millimeter_level_peg_in_hole}, light bulb removal \cite{2025_corl_flying_hand}, and in-flight grasping \cite{2024_npj_robotics_aerial_grasp}. Alternatively, manipulation can also be achieved through direct attachment or cable suspension, with cable-suspended payload transportation being a prominent example \cite{2024_tro_impact_aware_payloads, 2024_tro_human_aware_payload_transportation, 2025_arxiv_resilient_load_transportation}.

Despite these advances in aerial manipulation, significant challenges persist in real-world applications. Most existing systems are confined to direct physical contact, yet many practical scenarios require indirect manipulation, which involves interacting with objects via an intermediate tool or platform. For instance, during transportation, a robot might need to carry a box using a cable-suspended carrier. Here, the robot must manipulate the carrier rather than the box directly, while actively stabilizing it to prevent falling. Such forms of indirect manipulation introduce new challenges, especially in learning to control intermediary tools.

\begin{figure}[!htbp]
    \centering
    \includegraphics[width=0.9\linewidth]{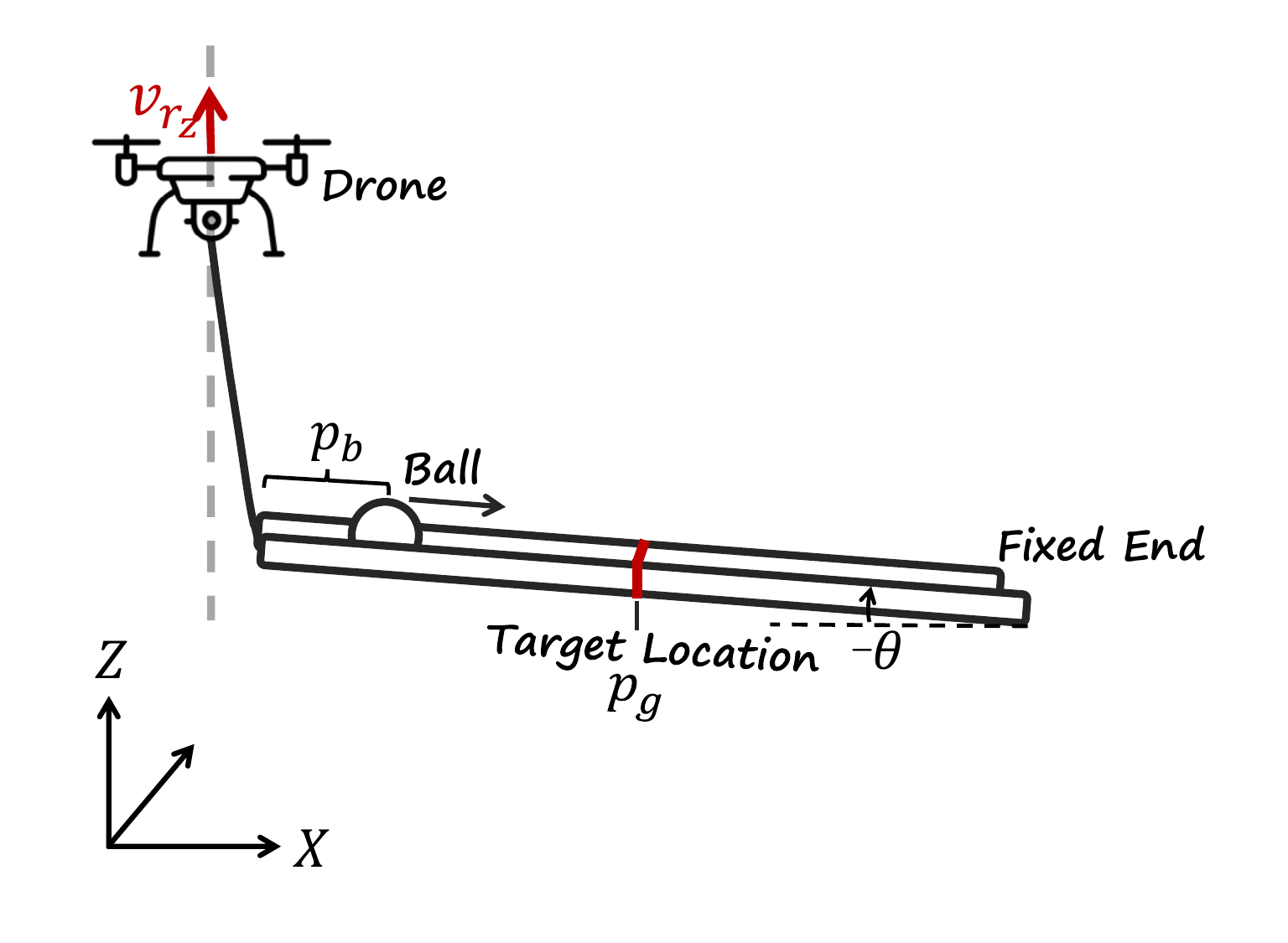}
    \caption{Illustration of the drone ball-balancing task. The drone is tethered to one end of the beam while the opposite end is fixed. The goal of the drone is to guide the ball toward the target position by manipulating the beam.}
    \label{drone_ball_balancing_problem}
\end{figure}

To explore the potential of indirect aerial manipulation, this paper studies a drone ball-balancing task (see Fig. \ref{drone_ball_balancing_problem}), where a drone tethered to one beam end via a rope and is required to manipulate a beam to position a ball at a target location. To solve this problem, we proposed a hierarchical control framework comprising: A high-level policy that generates velocity commands according to system state observations and a low-level velocity controller for precise reference tracking. 

Motivated by recent advances in RL \cite{2024_real_world_humaniod_locomotion, 2025_tro_extreme_adaptation}, we formulated the drone ball-balancing task as a Markov Decision Process (MDP) and developed an RL-based high-level controller to address this problem. Specifically, we created a drone ball-balancing simulator using Isaac Lab \cite{2023_RAL_isaac_lab}. To enable GPU-accelerated parallel training, we implemented a PyTorch-based low-level flight controller based on the SE(3) geometry controller \cite{2010_SE3_controller}. The balancing policy was parameterized by a neural network and trained using the Robust Policy Optimization algorithm (RPO) \cite{2022_rpo}. 

Simulation experiments demonstrate that the RL-based policy exhibits excellent performance in the drone ball-balancing task. To further validate the superiority of the RL-based approach, we replaced the high-level RL agent with incremental PID controllers \cite{2006_practical_pid_control} within the same hierarchical framework. Empirical results reveal that PID controllers fail to stabilize the system, even when augmented with additional velocity constraints.

While numerous studies demonstrate RL's superiority over classical control methods (e.g., PID and model predictive control (MPC)) in robotics \cite{2021_TRO_rl_for_robot_fish, 2022_RAS_rl_for_elastic_ligaments, 2022_RAL_adversary_rl}, these findings remain largely empirical, lacking systematic analysis of the fundamental factors contributing to RL's success or classical methods' limitations. In contrast, a recent comparative study in autonomous drone racing \cite{2023_science_robotics_rl_vs_mpc} reveals that RL's advantage stems primarily from its ability to optimize more effective objectives. This insight provides a principled foundation for future method selection and advancement in similar robotic applications. Inspired by this work, we conducted systematic controlled experiments to investigate why RL outperforms PID control. Our results demonstrate that RL's superiority stems not from discovering superior control gains or employing nonlinear error mapping, but rather from its ability to leverage richer state observations for decision-making effectively.

\section{Methodology}\label{methodology}
% This section systematically presents our methodology. First, we describe the hierarchical control framework designed for the human-drone collaborative ball-balancing task in Section \ref{framework}. Then, we detail the principles and designs of both RL-based high-level controller (Section \ref{rl_controller}) and PID-based high-level controller (Section \ref{pid_controller}).

\subsection{Hierarchical Framework}\label{framework}
% Although the drone's vertical velocity serves as the system's control input, we must additionally account for motion control of the drone itself due to its intrinsic instability. 
% Hierarchical control frameworks have proven highly effective for complex control tasks by decomposing them into manageable subproblems \cite{2021_RAL_planning_augmented_hierarchical_RL, 2022_TRO_RL_MPC_drone}. Building on this principle, 
% Based on this requirement, we proposed a two-layer control architecture as illustrated in Fig. \ref{hierarchical_framework}. 

\begin{figure}[!htbp]
    \centering
    \includegraphics[width=0.9\linewidth]{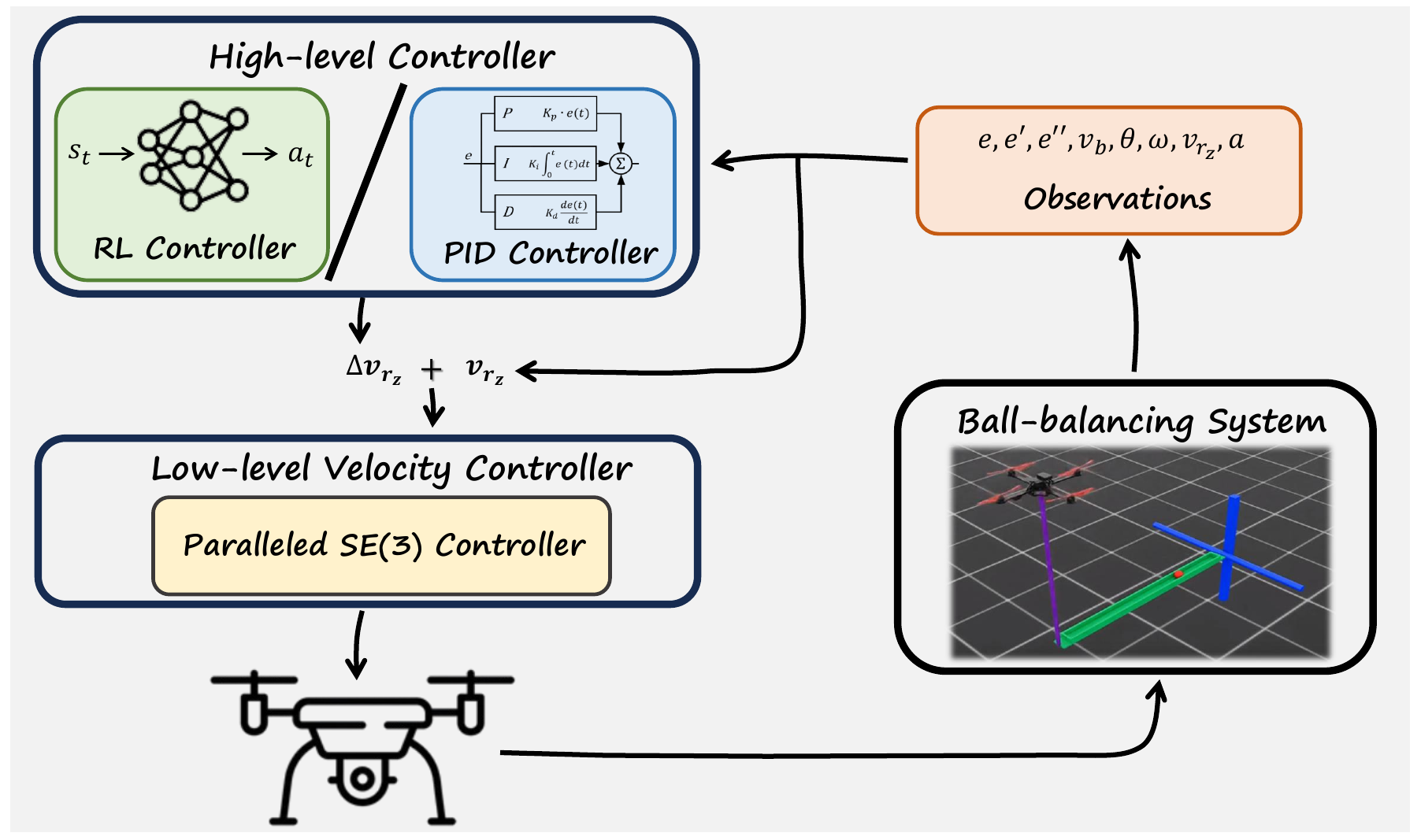}
    \caption{Illustration of the proposed hierarchical control framework. The high-level controller generates optimal Z-axis velocity commands based on real-time system state observations, while the low-level controller ensures precise tracking of these commanded velocities.}
    \label{hierarchical_framework}
\end{figure}

We proposed a two-layer control architecture as illustrated in Fig. \ref{hierarchical_framework} to solve the drone ball-balancing task. 

The high-level controller generates strategic commands for beam inclination adjustment. Its operational workflow is: receiving real-time system state observations; computing an appropriate vertical velocity increment ($\Delta v_{r_z}$) based on these observations; outputting this increment to be combined with the current drone velocity ($v_{r_z}$), forming the reference velocity ($v_{r_z}+\Delta v_{r_z}$). The high-level controller generates velocity increments rather than direct velocity commands to account for the drone's physical constraints. This approach ensures all commanded velocities remain within the drone's achievable acceleration limits, preventing the generation of kinematically infeasible reference velocities.

The low-level controller executes precise velocity tracking for the drone. It takes the reference velocity from the high-level controller and computes the corresponding rotor speeds for flight control. In this paper, we implemented a PyTorch-based controller based on the SE(3) geometry controller \cite{2010_SE3_controller} to enable GPU-accelerated parallel training.
% Given the technological maturity of current drone flight controllers, we integrated proven control solutions rather than developing custom implementations. For simulation, we implemented a PyTorch-based controller based on the SE(3) geometry controller \cite{2010_SE3_controller} to enable GPU-accelerated parallel training. 

% For real-world deployment, we employed the PX4 Autopilot as the low-level flight controller. A detailed discussion of the low-level controllers is omitted here as it falls outside the paper's primary research objectives.

\subsection{RL-based High-level Controller}\label{rl_controller}

A common practice in applying RL to robotics is to define the state, action, and reward function for the MDP. In the drone ball-balancing task, the drone's observations consist of the state information of the ball-and-beam system and its own current state, represented as an 8-dimensional state vector: $[e, e', e'', v_b, \theta, \omega, v_{r_z}, a]$. Here, $e$ denotes the position error between the ball's current position $p_b$ and the goal position $p_g$; $e'$ is the first-order difference of the position error, calculated as $e' = e_t - e_{t-1}$; $e''$ is the second-order difference, computed as $e'' = (e_t - e_{t-1}) - (e_{t-1} - e_{t-2})$; $v_b$ represents the ball's velocity along the beam; $\theta$ is the beam angle; $\omega$ is the angular velocity of the beam; and $a$ denotes the last action taken by the agent, which is the one-dimensional vertical velocity increment $\Delta v_{r_z}$. To formalize the control objective, we design a composite reward function as follows:
\begin{equation}
    r = r_{\text{object}} + r_{\text{control}} + r_{\text{failure}} + r_{\text{goal}},
\end{equation}
% where $r_{\text{object}}$ penalizes the ball's position error $e$ and velocity $v_b$, $r_{\text{control}}$ regularizes the drone's vertical velocity $v_{r_z}$ and the taken action $a$, $r_{\text{failure}}$ is a binary penalty term that triggers when the system enters a terminal state, defined as either: exceeding maximum beam tilt (i.e., $|\theta| > \theta_{max}$) or ball detachment from the beam (i.e., $|e| > e_{\max}$), and $r_{\text{goal}}$ encourages progress towards the target position when the ball enters a target neighborhood (i.e., $|e|<e_{goal}$). The reward components are mathematically defined as:
where its components are mathematically defined as:
\begin{subequations}
    \begin{equation}
    r_{\text{object}} = -k_1 e^2 - k_2 v_b^2, 
    \end{equation}
    \begin{equation}
        r_{\text{control}} = -k_3 v_{r_z}^2 - k_4 a^2, 
    \end{equation}
    \begin{equation}
        r_{\text{failure}} = \begin{cases}
        -k_5 & \text{if } (|\theta| > \theta_{\max}) \lor (|e| > e_{\max}) \\
        0 & \text{otherwise}
        \end{cases},
    \end{equation}
    \begin{equation}
        r_{\text{goal}} = \begin{cases}
        (c - \frac{c - c_{\min}}{e_{\text{goal}}} |e|) \times \exp{(-k_6 |v_b|)} & \text{if } |e| < e_{\text{goal}} \\
        0 & \text{otherwise}
        \end{cases}.
    \end{equation}
\end{subequations}
Given the outstanding performance of RPO \cite{2022_rpo}, it was selected to train the ball-balancing policy.

% Proximal Policy Optimization (PPO) \cite{2017_ppo} has become a dominant algorithm for robotic control tasks. For continuous action spaces, PPO typically parameterizes actions as Gaussian distributions, where the trainable mean and standard deviation govern exploitation and exploration, respectively. While effective, PPO policies often exhibit decreasing entropy during training, corresponding to reduced exploration. To address this limitation, Robust Policy Optimization (RPO) \cite{2022_rpo} was developed to sustain higher exploration levels throughout training. Empirical results across standard continuous control benchmarks demonstrate RPO's advantages over PPO in maintaining policy entropy and enhancing performance. Therefore, this paper selected RPO to train the ball-balancing policy\footnote{Experimental comparisons between RPO and PPO in the ball-balancing task confirmed RPO's better performance, though these results are omitted for brevity.}.

% RPO combines Gaussian and Uniform distributions to construct a new action distribution. Specifically, at each training timestep, the mean $\mu$ of Gaussian distribution $\mathcal{N(\mu,\sigma)}$ is perturbed by adding a random shift $z$ sampled from Uniform distribution $\mathcal{U}(-\alpha, \alpha)$, resulting in a new action distribution $\mathcal{N}(\mu', \sigma)$ with shifted mean $\mu' = \mu + z$. Consequently, actions are sampled from the modified distribution $\mathcal{N}(\mu', \sigma)$.

\subsection{PID-based High-level Controller}\label{pid_controller}
While PID control has demonstrated success in traditional ball-and-beam systems \cite{2015_pid_ball_on_beam}, its efficacy in the drone ball-balancing task remains unexplored. This work adopts an incremental PID formulation aligned with the design where the high-level controller outputs vertical velocity increments. The discrete-time implementation of this incremental PID controller follows:
\begin{equation}\label{incremental_pid}
        a_{t_k} = K_p e_{t_k}' + \frac{K_p \Delta t}{T_i} e_{t_k} + \frac{K_p T_d}{\Delta t} e_{t_k}'',
\end{equation}
where $a_{t_k}$ denotes the drone's high-level commanded action at timestep $t_k$, $\Delta t$ represents the sampling time interval, $K_p$ is the proportional gain, $T_i$ and $T_d$ are the integral and derivative time constants, respectively. Through extensive empirical tuning, we consistently observed system oscillations regardless of gain selection. However, these oscillations attenuated under vertical velocity constraints. This finding motivated our comparison with three distinct velocity constraint conditions: strict constraint (i.e., $|v_{r_z}| \leq 0.1\text{m/s}$), moderate constraint (i.e., $|v_{r_z}| \leq 0.3 \text{m/s}$), and loose constraint (i.e., $|v_{r_z}| \leq 0.5\text{m/s}$). For each condition, we empirically determined optimal PID gains to enable a comprehensive performance comparison with the RL controller.
\section{Experiments}\label{experiments}
% This section designs both simulation and real-world experiments to answer the following research questions:
% \begin{itemize}
% \item \textit{Q1}: Can the task-optimized RL-based high-level controller maintain robustness against goal variations and human disturbances?
% \item \textit{Q2}: How does the RL-based controller compare with the PID-based controller in performance?
% \item \textit{Q3}: Can the trained RL-based high-level controller in simulation successfully transfer to real-world human-drone ball-balancing scenarios?
% \end{itemize}
% Specifically, we present the experimental setups in Section \ref{setups}. Then, Section \ref{simulation_evaluation} answers \textit{Q1} and \textit{Q2} through simulation evaluations, while Section \ref{real_world_test} addresses \textit{Q3} via real-world validation.

% \begin{figure}[!htbp]
%     \centering
%     \includegraphics[width=0.9\linewidth]{figures/simulation_scene.png}
%     \caption{Screenshot of the drone ball-balancing task in simulation.}
%     \label{simulation_scene}
% \end{figure}

\subsection{Setups}\label{setups}
To meet the data requirements for RL policy training, we developed a drone ball-balancing simulator using the Isaac Lab framework \cite{2023_RAL_isaac_lab}. As shown in Fig. \ref{hierarchical_framework} (Ball-balancing System), one end of the beam is fixed at a constant altitude with a rotatable joint supporting the rotation of the beam, while the drone controls the tethered
opposite end. The simulation uses Ascending Technologies’ Hummingbird drone model.

For training the RL-based high-level policy, we designed both actor and critic networks as two-layer MLPs with 256 neurons per layer, where the actor's output layer employed a \textit{tanh} activation function. Each training episode lasted up to 5 seconds, during which the drone needed to quickly position and stabilize the ball at the target. To mitigate instability issues during initial training phases, we implemented early termination when either the beam tilt exceeded $\theta_{\max}$ or the ball detached from the beam. This mechanism significantly enhanced exploration efficiency. The control architecture operated with high-level policy updates at 60 Hz and low-level velocity tracking at 180 Hz to ensure stable and responsive control. 

\subsection{Results}\label{simulation_evaluation}
\subsubsection{Evaluation Metrics}
The evaluation protocol runs 1000 test episodes, each lasting 10 seconds. For the RL-based controller, we report performance as means and standard deviations across five random seeds. This paper utilizes four evaluation metrics: success rate (SR), steady-state error (SE), convergence time (CONT), and climb time (CLIT). Specifically, SR calculates the percentage of episodes with absolute terminal position error below $0.01\text{m}$; SE measures the absolute terminal position error; CONT tracks the time required for the ball to reach and stay within $\pm0.02\text{m}$ of the target; and CLIT captures the initial entry time into the target zone with position error $0.02\text{m}$. 
% and FR tallies episodes where either the beam tilt exceeds $\theta_{\max}$ or the absolute ball's position error surpasses $e_{\max}$. 

\begin{table}[!htbp]
	\caption{Performance comparison between RL and PID Controllers.}
	\label{static_results_0.5}
	\centering
            \scalebox{0.7}{
		\begin{tabular}{c|c|c|c|c}
			\toprule
			\toprule
			\textbf{Controller} & \textbf{SR}($\%$)$^{\uparrow}$ & \textbf{SE}(mm)$^{\downarrow}$ & \textbf{CONT}(s)$^{\downarrow}$ & \textbf{CLIT}(s)$^{\downarrow}$ \\
			\midrule
			  \textbf{RL} & \boldmath{$100\pm0$} & \boldmath{$1.290\pm0.535$} & \boldmath{$1.702\pm0.117$} & $1.383\pm0.027$ \\
			\textbf{PID with} \boldmath{$|v_{r_z}| \leq 0.1$} & \boldmath{$100$} & $1.919$ & $2.623$ & $2.622$ \\
			\textbf{PID with} \boldmath{$|v_{r_z}| \leq 0.3$} & $61$ & $9.548$ & $3.508$ & $1.107$ \\
                \textbf{PID with} \boldmath{$|v_{r_z}| \leq 0.5$} & $14$ & $30.275$ & $9.977$ & \boldmath{$0.952$} \\
			\bottomrule
			\bottomrule
		\end{tabular}}
\end{table}
\begin{figure}[!htbp]
    \centering
        \begin{minipage}{0.48\linewidth}
		\centerline{\includegraphics[width=0.23\paperwidth]{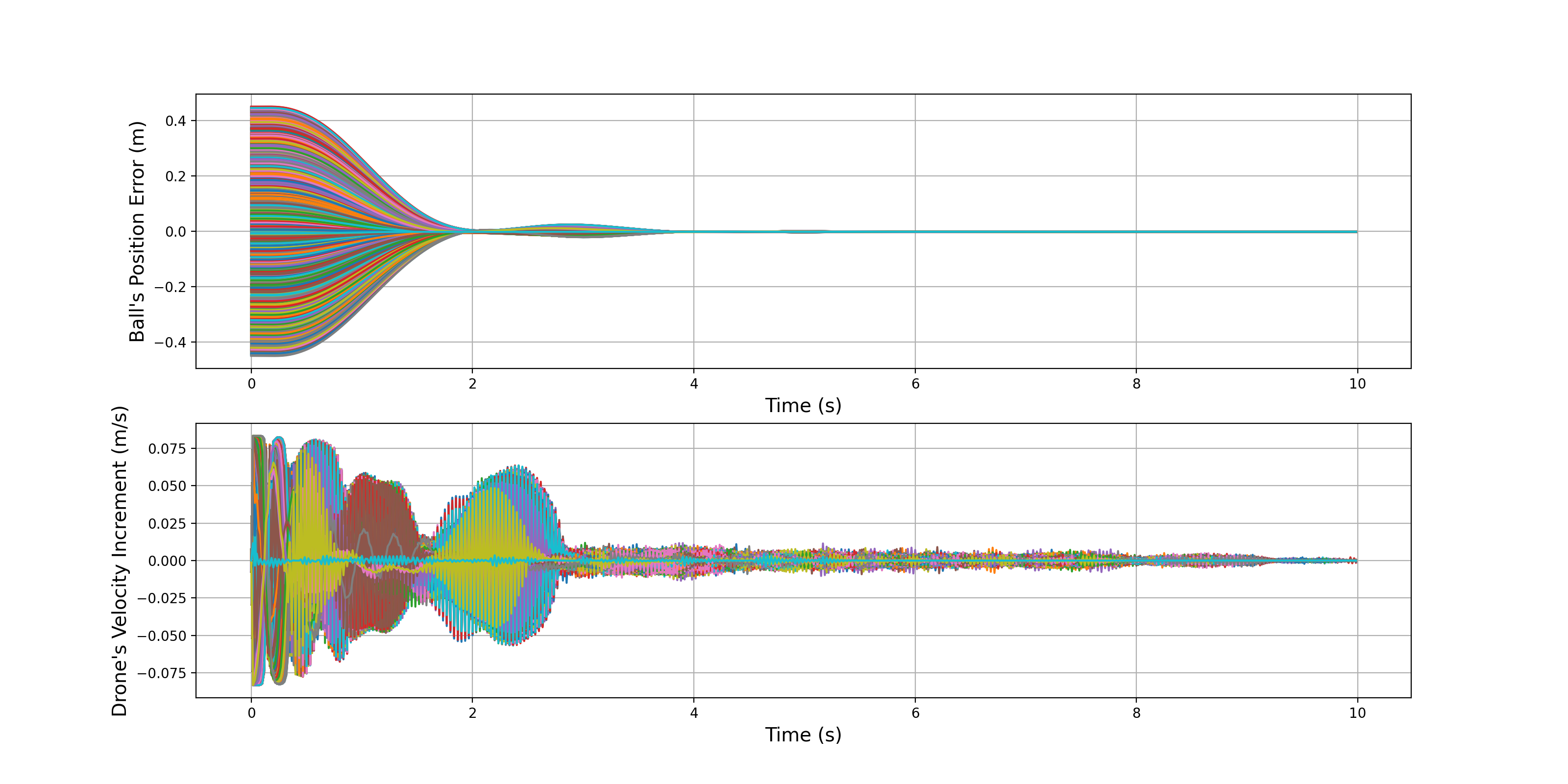}}
		\centerline{\scriptsize (a) RL-based controller.}
	\end{minipage}
        \hspace{0.1cm}
	\begin{minipage}{0.48\linewidth}
		\centerline{\includegraphics[width=0.23\paperwidth]{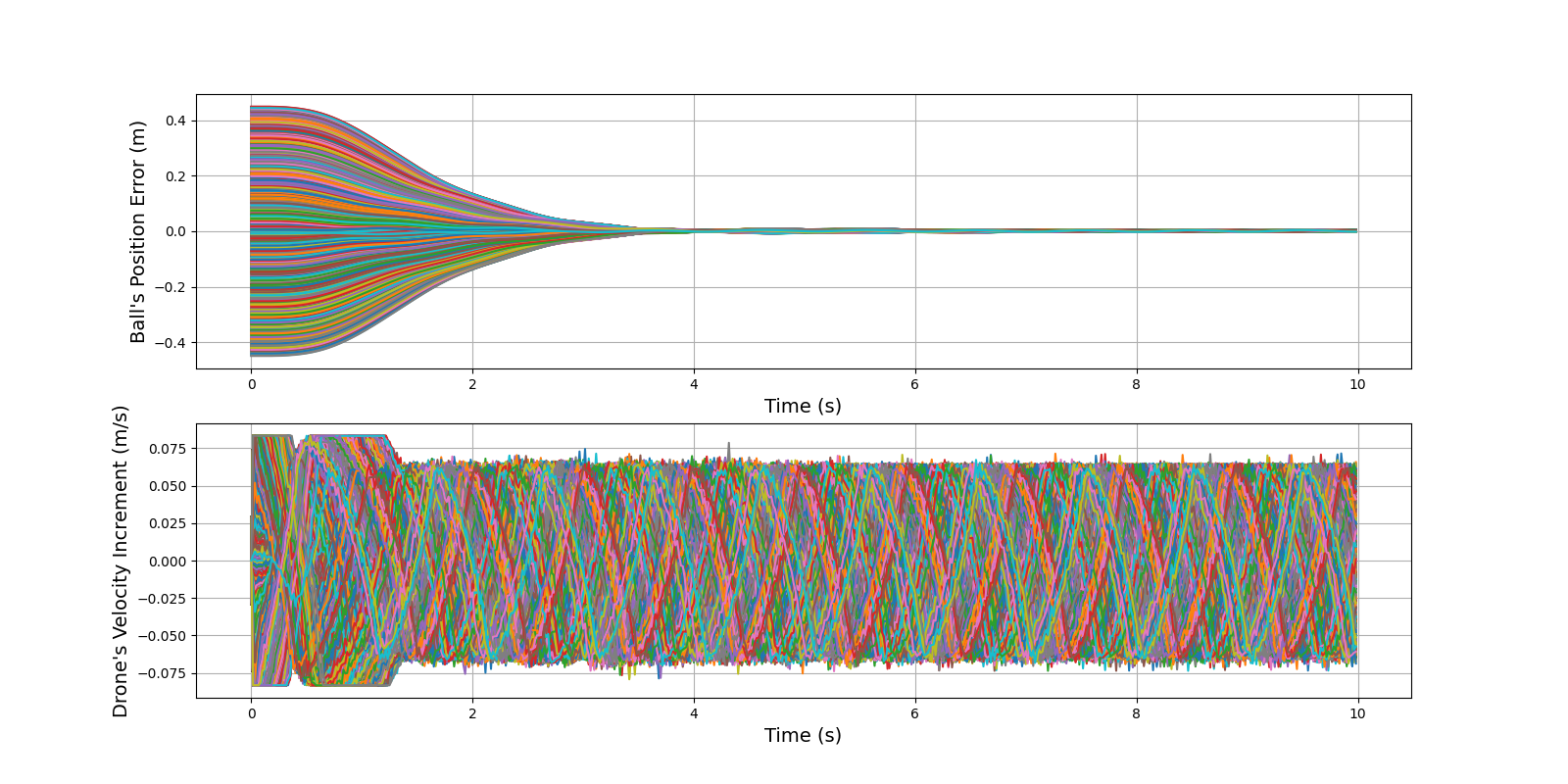}}
		\centerline{\scriptsize (b) PID controller with $|v_{r_z}| \leq 0.1$.}
	\end{minipage}
	% \vfill
	\begin{minipage}{0.48\linewidth}
		\centerline{\includegraphics[width=0.23\paperwidth]{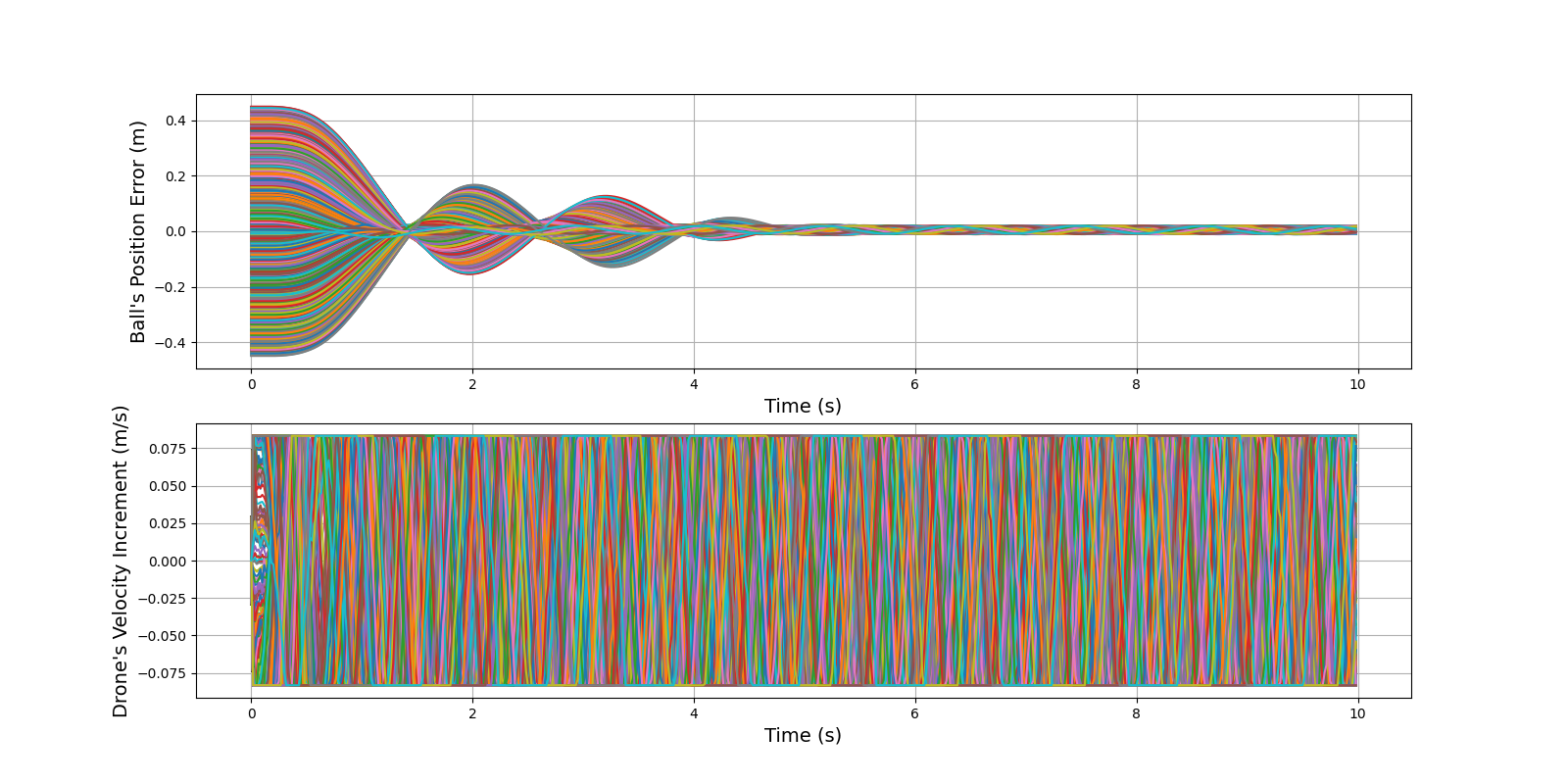}}
		\centerline{\scriptsize (c) PID controller with $|v_{r_z}| \leq 0.3$.}
	\end{minipage}
        \hspace{0.1cm}
	\begin{minipage}{0.48\linewidth}
		\centerline{\includegraphics[width=0.23\paperwidth]{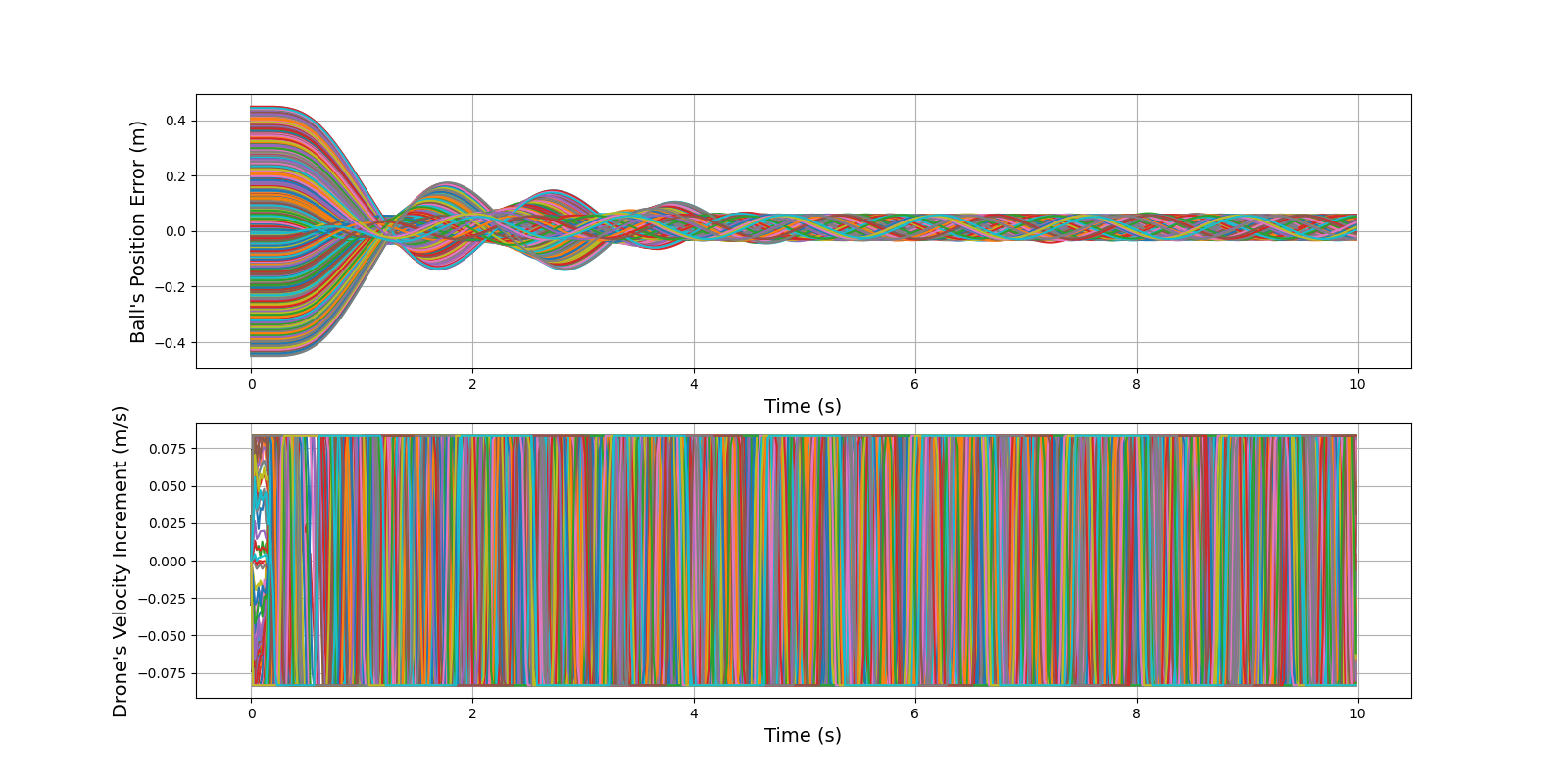}}
		\centerline{\scriptsize (d) PID controller with $|v_{r_z}| \leq 0.5$.}
        \end{minipage}
    \caption{Time evolution of the ball's position error and drone's vertical velocity increment for the drone ball-balancing task with $p_g=0.5$. (a) demonstrates the RL-based controller's fast and stable performance, while (b)-(d) reveal oscillatory behavior in the PID controllers with velocity constraints, despite the position errors seeming to converge to the surrounding region of the target at (b) and (c).}\label{curve_rl_vs_pid_static_goal_0.5}
\end{figure}

\subsubsection{Performance}
 The quantitative results, including the ball’s position error and the drone’s vertical velocity increment over time, are presented in Table \ref{static_results_0.5} and Fig. \ref{curve_rl_vs_pid_static_goal_0.5}, respectively. As shown in Table \ref{static_results_0.5}, the RL-based controller achieves a $100\%$ SR and outperforms all PID-based controllers in both SE and CONT. For the PID controllers with velocity constraints, the SR decreases as the constraints are relaxed, highlighting PID’s limitations in this task. Interestingly, while the PID controller with $|v_{r_z}| \leq 0.1$ achieves comparable performance to the RL-based method, its control actions (i.e., drone velocity increments $\Delta v_{r_z}$) exhibit high-frequency oscillations (see Fig. \ref{curve_rl_vs_pid_static_goal_0.5} (b)). This indicates that the system oscillates around the target position, a behavior that becomes more pronounced with looser velocity constraints (see Fig. \ref{curve_rl_vs_pid_static_goal_0.5} (c) and (d)).

\section{Why RL Outperforms PID}\label{rl_vs_pid}
% In this section, we empirically investigate why RL outperforms PID control. Building upon the three hypotheses established in Section \ref{hypotheses}, we designed controlled experiments detailed in Section \ref{experiment_design} and presented the results in Section \ref{empirical_results}.

\subsection{Hypotheses}\label{hypotheses}
We empirically propose three hypotheses to explain the performance differences between RL and PID control:
\begin{itemize}
\item \textit{H1}: RL's gradient-based optimization enables discovery of more effective control parameters than PID's manual trial-and-error tuning.
\item \textit{H2}: RL's nonlinear policy mapping better handles the system's nonlinear dynamics compared to PID's linear control law.
\item \textit{H3}: RL's state representation, incorporating richer system information beyond position error, facilitates superior control decisions versus PID.
\end{itemize}

\subsection{Experiment Design}\label{experiment_design}
Designing targeted experiments to validate each hypothesis presents significant methodological challenges. As an alternative approach, we evaluate RL performance using identical input features to the PID controller for direct comparison. Following Equation (\ref{incremental_pid}), we provide the RL actor with the same three-dimensional feedback ($e, e', e''$) used by the PID controllers. Simultaneously, we constrain the vertical velocity to make consistent comparison conditions with the PID controllers. However, to maintain learning efficiency, we implement a privileged critic architecture that retains access to the full 8-dimensional state space $[e, e', e'', v_b, \theta, \omega, v_{r_z}, a]$ for state-action value estimation. To enhance training effectiveness, we developed a three-phase curriculum learning strategy:
\begin{itemize}
    \item \textbf{Initial Phase}: Target neighborhood $e_{\text{goal}} = 0.3$ with ball initialization outside this region.
    \item \textbf{Intermediate Phase}: Reduced target neighborhood ($e_{\text{goal}} = 0.15$).
    \item \textbf{Final Phase}: Tight convergence requirement ($e_{\text{goal}} = 0.05$).
\end{itemize}
This progressive approach ensures comprehensive coverage of the initial state space while gradually increasing task difficulty.

\subsection{Empirical Results}\label{empirical_results}
\begin{figure}[!htbp]
    \centering
        \begin{minipage}{0.48\linewidth}
		\centerline{\includegraphics[width=0.23\paperwidth]{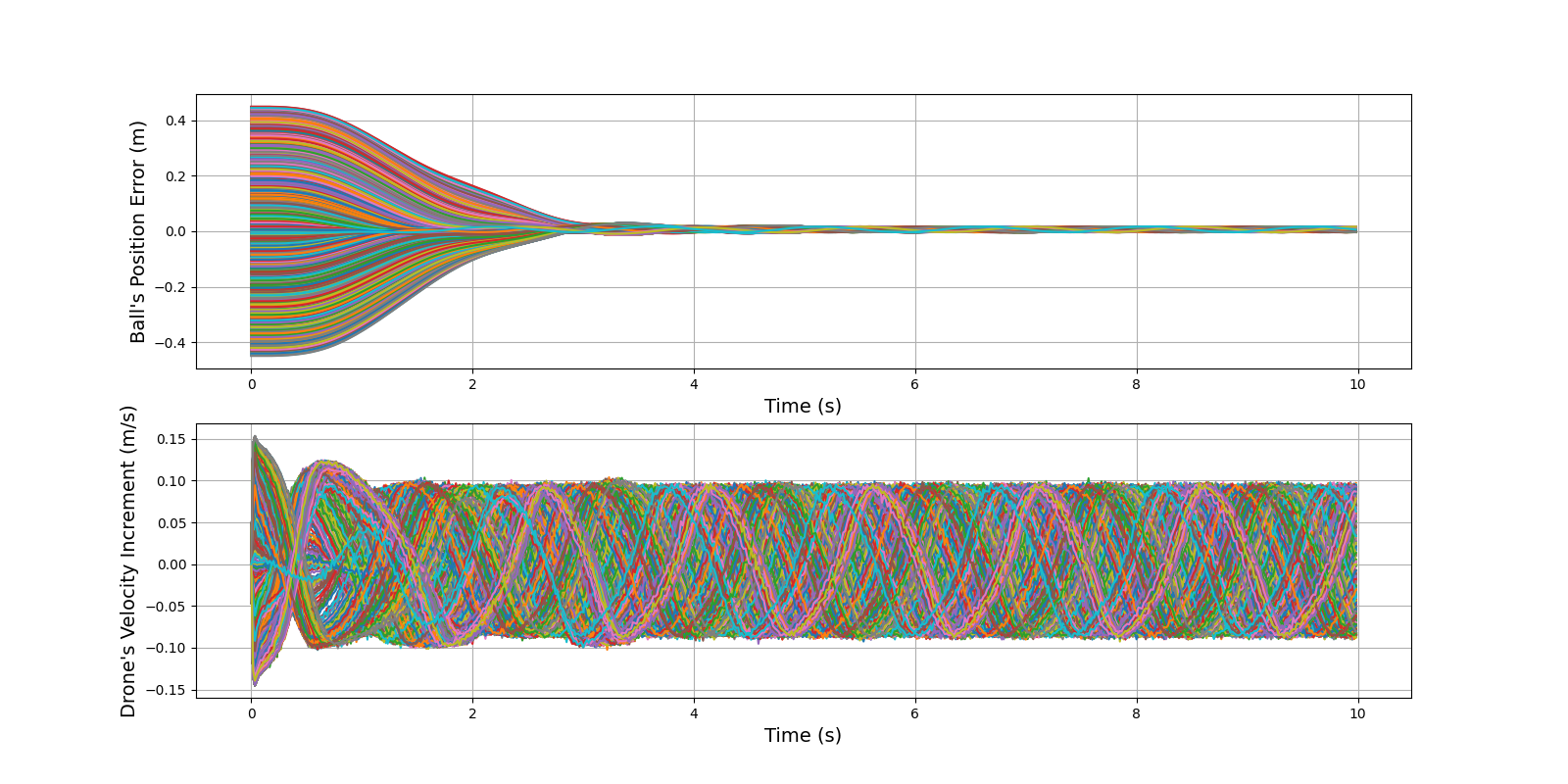}}
		\centerline{\scriptsize (a) RL controller with $|v_{r_z}| \leq 0.1$.}
	\end{minipage}
        \hspace{0.1cm}
	\begin{minipage}{0.48\linewidth}
		\centerline{\includegraphics[width=0.23\paperwidth]{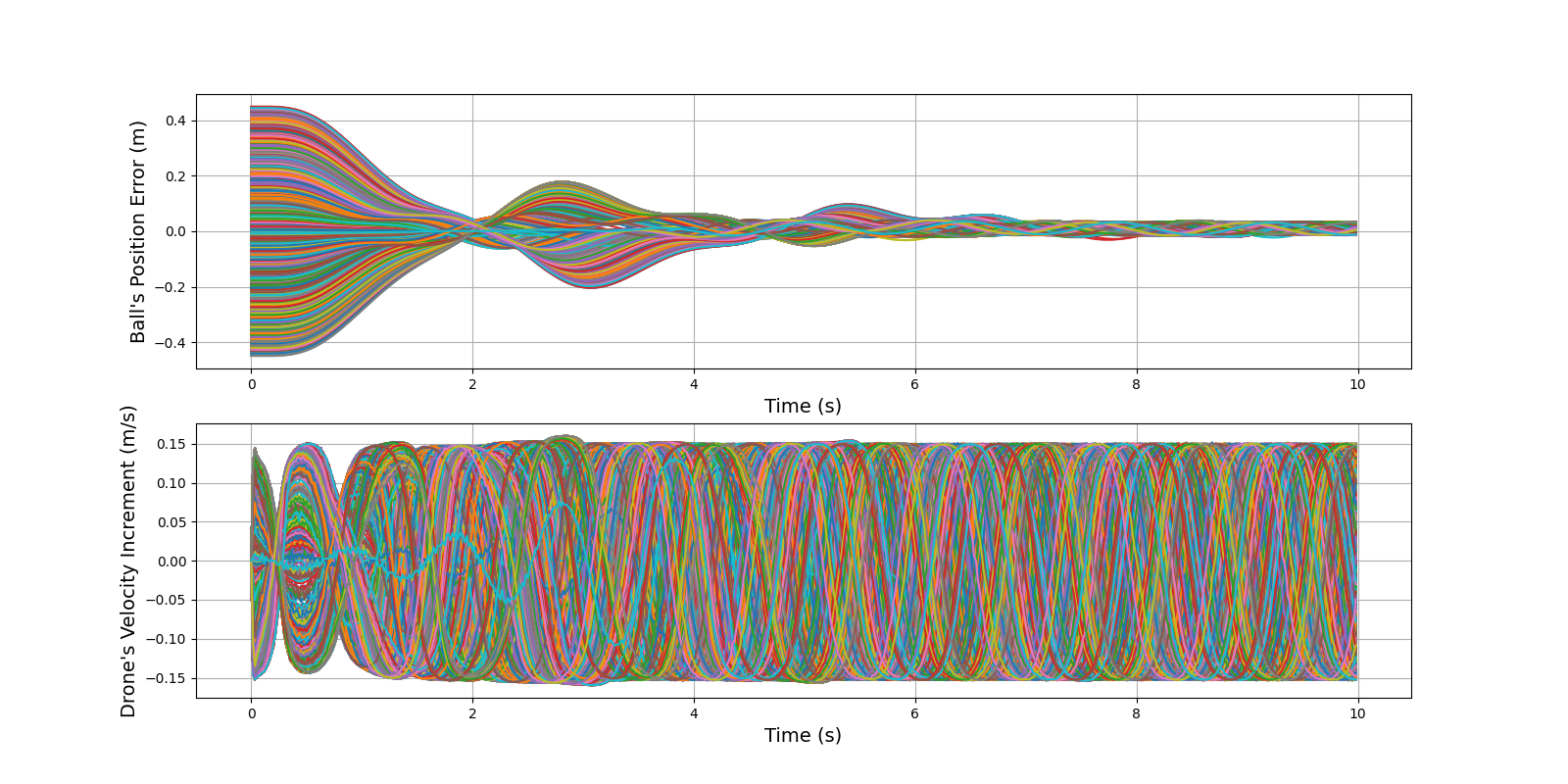}}
		\centerline{\scriptsize (b) RL controller with $|v_{r_z}| \leq 0.3$.}
	\end{minipage}
        % \vfill
	\begin{minipage}{0.48\linewidth}
		\centerline{\includegraphics[width=0.23\paperwidth]{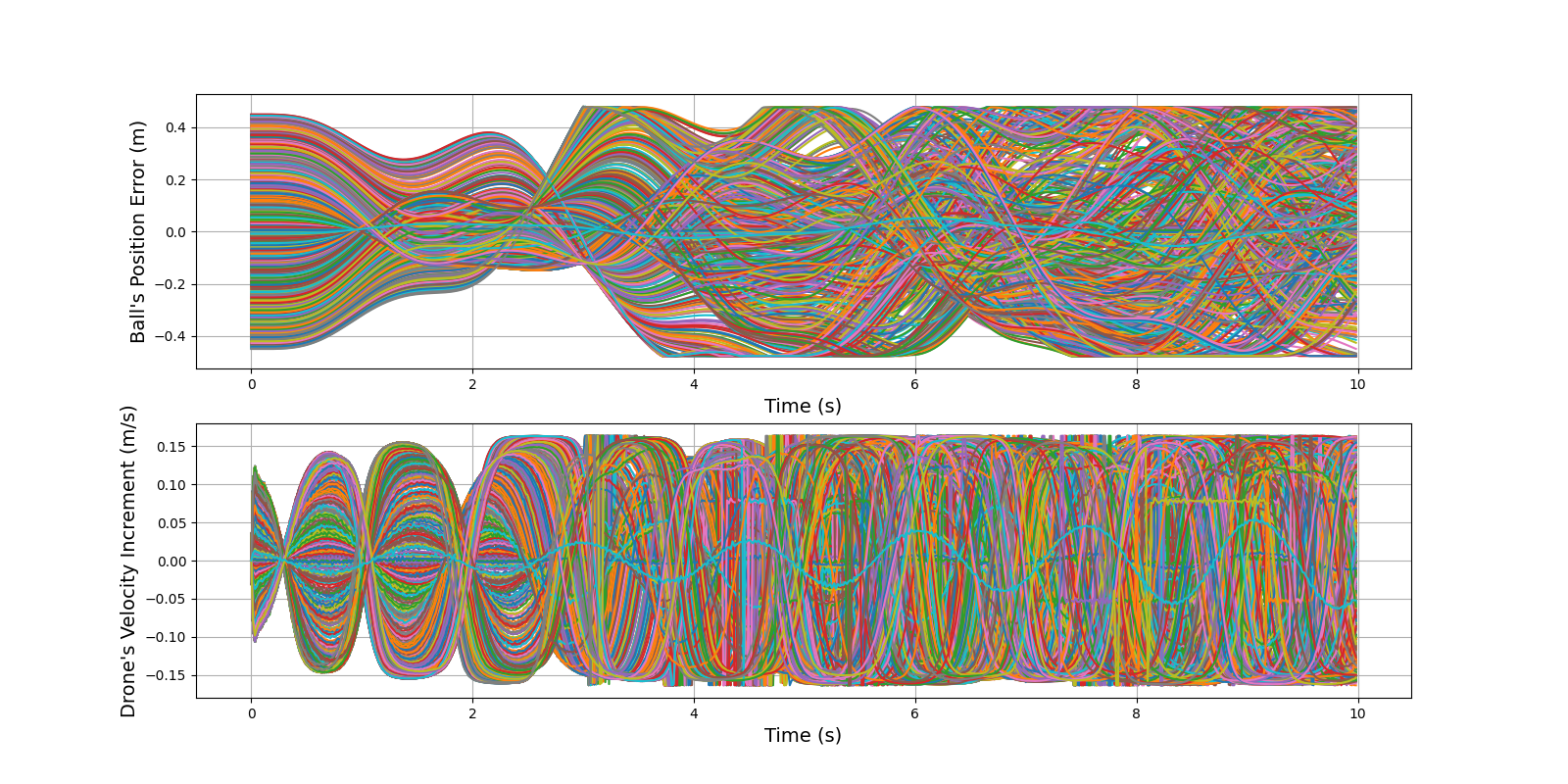}}
		\centerline{\scriptsize (c) RL controller with $|v_{r_z}| \leq 0.5$.}
	\end{minipage}
        \hspace{0.1cm}
        \begin{minipage}{0.48\linewidth}
    		\centerline{\includegraphics[width=0.23\paperwidth]{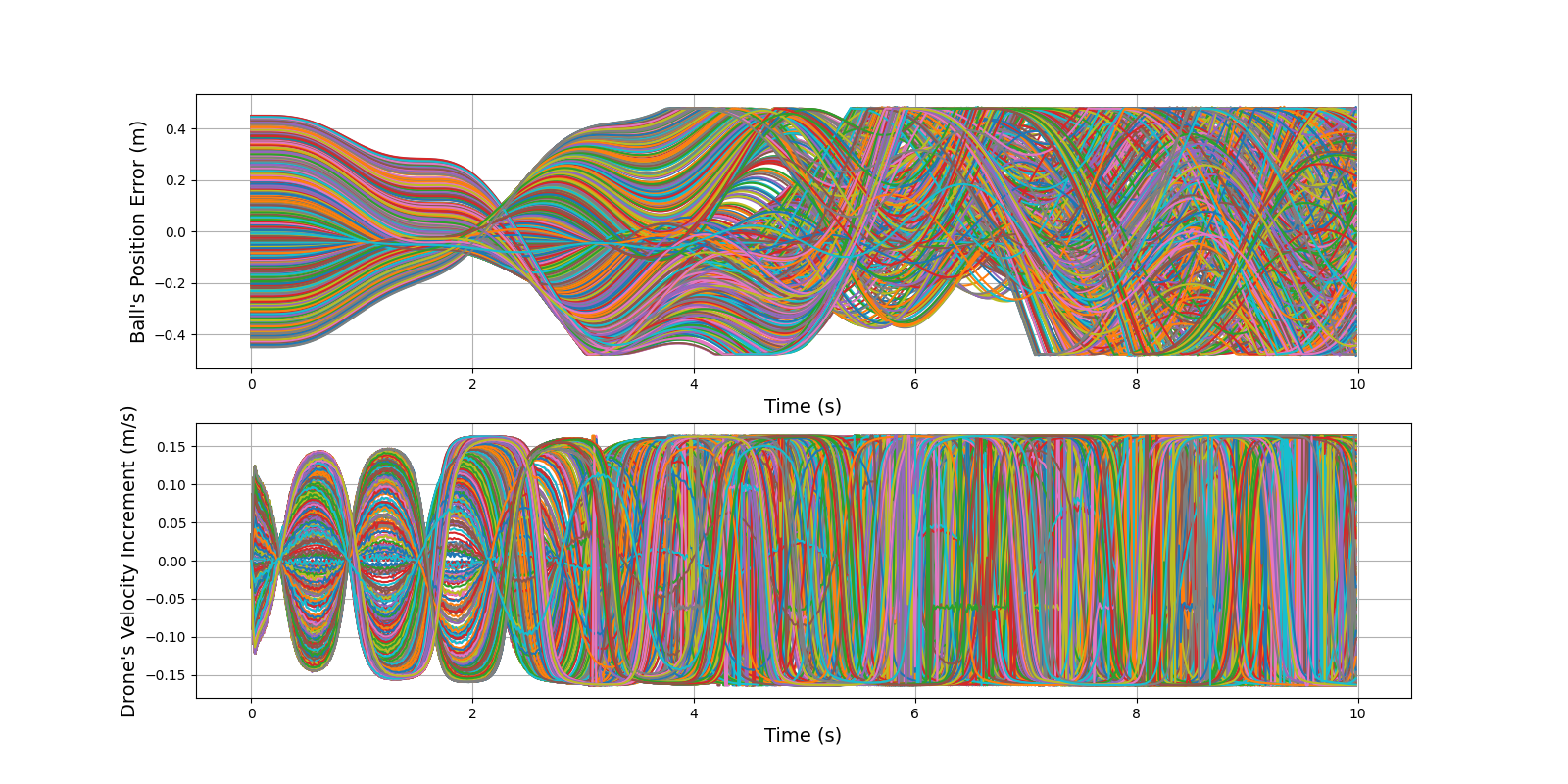}}
    		\centerline{\scriptsize (d) RL controller without Constraint.}
	\end{minipage}
    \caption{Time evolution of the ball's position error and drone's vertical velocity increment for RL using PID-equivalent inputs in the drone ball-balancing task with $p_g=0.5$.}\label{curve_rl_vs_pid_vel_constraint_static_goal_0.5}
\end{figure}

We trained RL policies using PID-equivalent inputs for the drone ball-balancing task with $p_g=0.5$ and evaluated their performance under identical conditions. As shown in Table \ref{RL_vs_PID_static_results_0.5} and Fig. \ref{curve_rl_vs_pid_vel_constraint_static_goal_0.5}, the RL controllers exhibit poorer performance compared to PID with velocity constraints, while also displaying characteristic oscillations around the target position.

This poor performance under identical inputs refutes hypotheses H1 (superior parameter optimization) and H2 (nonlinear mapping advantage), indicating that RL’s performance benefits do not originate from these factors. However, when provided with full state feedback, RL demonstrates significantly improved performance. This strongly supports hypothesis H3, confirming that RL’s advantage derives from its ability to leverage richer system information for more effective decision-making. This result highlights the potential for improving the controller’s performance by deploying more sensors.
\begin{table}[!htbp]
	\caption{Performance of RL using PID-equivalent inputs with velocity constraint conditions.}
	\label{RL_vs_PID_static_results_0.5}
	\centering
            \scalebox{0.7}{
		\begin{tabular}{c|c|c|c|c}
			\toprule
			\toprule
			\textbf{Controller} & \textbf{SR}($\%$)$^{\uparrow}$ & \textbf{SE}(mm)$^{\downarrow}$ & \textbf{CONT}(s)$^{\downarrow}$ & \textbf{CLIT}(s)$^{\downarrow}$ \\
			\midrule
			  \textbf{RL with} \boldmath{$|v_{r_z}| \leq 0.1$} & $90.360\pm17.960$ & $5.709\pm1.566$ & $2.038\pm0.342$ & $1.869\pm0.084$ \\
                \textbf{RL with} \boldmath{$|v_{r_z}| \leq 0.3$} & $36.040\pm14.702$ & $20.447\pm11.361$ & $9.697\pm0.321$ & $1.709\pm0.060$ \\
                \textbf{RL with} \boldmath{$|v_{r_z}| \leq 0.5$} & $7.120\pm2.881$ & $117.454\pm41.578$ & $9.966\pm0.035$ & $1.617\pm0.097$ \\
                \textbf{RL without Constraint} & $2.120\pm0.801$ & $257.910\pm33.050$ & $10.011\pm0.004$ & $1.733\pm0.192$ \\
   %              \midrule
			% \multirow{3}{*}{\textbf{PID}} & \boldmath{$v_{r_z} \leq 0.1$} & $100$ & $2.270$ & $2.627$ & $2.624$ & $0$ \\
			% ~ & \boldmath{$v_{r_z} \leq 0.3$} & $62$ & $14.332$ & $9.770$ & $2.353$ & $0$ \\
   %              ~ & \boldmath{$v_{r_z} \leq 0.5$} & $0$ & $0$ & $0$ & $0$ & $0$ \\
			\bottomrule
			\bottomrule
		\end{tabular}}
\end{table}

%%%%%%%%%%%%%%%%%%%%%%%%%%%%%%%%%%%%%%%%%%%%%%%%%%%%%%%%%%%%%%%%%%%%%%%%%%%%%%%%
\section{Conclusions}\label{conclusions}
In this work, we investigate the potential of model-free RL for indirect aerial manipulation by proposing a hierarchical control framework with an RL-based high-level controller for a drone ball-balancing task. Simulated results demonstrate that the RL policy achieves better performance than PID control. Systematic analysis shows that this advantage stems primarily from RL's ability to exploit richer state information rather than from superior control parameter optimization or inherent nonlinear mapping capabilities. These findings underscore the critical role of comprehensive state representation in learning-based control systems and highlight the potential for improving controller performance through the deployment of additional sensors.0

% \section*{APPENDIX}
% \section*{ACKNOWLEDGMENT}

\bibliographystyle{IEEEtran}
\bibliography{root}

\end{document}